\pdfoutput=1
\documentclass[11pt]{article}
\usepackage[preprint]{acl}
\usepackage{times}
\usepackage{latexsym}
\usepackage[T1]{fontenc}
\usepackage[utf8]{inputenc}
\usepackage{microtype}
\usepackage{graphicx}
\usepackage{booktabs}
\usepackage{amsmath,amssymb}
\usepackage{xcolor}
\usepackage{colortbl}
\definecolor{oursl}{HTML}{FBEDEB}
\newcommand{\drop}[1]{\textcolor[HTML]{C53030}{\textbf{#1}}}
\usepackage{caption}
\newcommand{\oursname}{\textsc{Lookout}}
\newcommand{\icon}[1]{\raisebox{-0.18em}{\includegraphics[height=0.95em]{figs/icon_#1.pdf}}}

\title{Rethinking Task Success for Embodied LLM Agents}

\author{Zhimei Chen\textsuperscript{*} \\
  Independent Researcher \\\And
  Mu Chen\textsuperscript{*}\textsuperscript{\dag} \\
  Mohamed bin Zayed University \\
  of Artificial Intelligence \\
  \texttt{mu.chen@mbzuai.ac.ae}}

\hypersetup{pdftitle={Rethinking Task Success for Embodied LLM Agents},pdfauthor={Zhimei Chen, Mu Chen}}

\begin{document}
\makeatletter
\g@addto@macro\@maketitle{%
\begin{center}
\includegraphics[width=\textwidth]{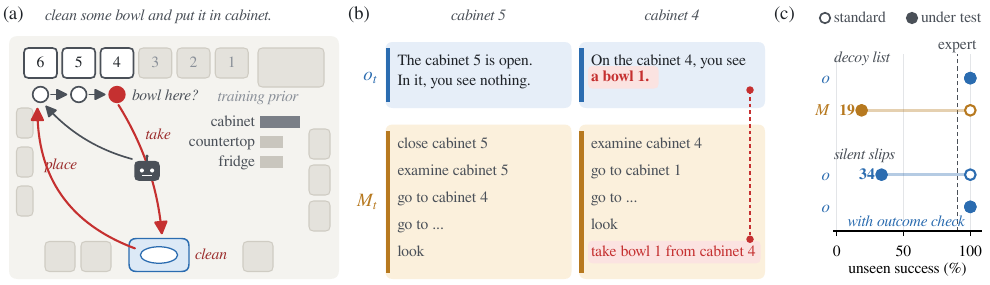}
\captionof{figure}{ALFWorld can be solved by a fixed search order that reads one fact per stop, and the action list supplies that fact. (a) A script checks for a bowl at each stop. (b) Both inputs show the bowl. (c) Unseen success.}
\label{fig:teaser}
\end{center}
\vspace{0.1in}}
\makeatother
\maketitle
{\renewcommand{\thefootnote}{}\footnotetext{\raggedright\textsuperscript{*}Equal contribution (Zhimei Chen and Mu Chen).\\\textsuperscript{\dag}Corresponding author: Mu Chen.}}

\begin{abstract}
Trained language agents complete about 90\% of the household tasks in ALFWorld, a benchmark of embodied text games. Top-leading solutions attribute these gains to better trajectory data and credit assignment, and read the success rate as evidence that the agent acts on what it observes. However, we find that ALFWorld can be solved with almost no observation: a script that visits receptacles in an order fixed by training-set frequencies and checks one fact per stop, whether the target is present, solves all unseen games. More importantly, even this fact need not be observed, since the list of admissible actions that trained agents receive at every step contains it. Success rate thus cannot distinguish an agent that perceives from one that follows a routine. We introduce \oursname{}, an evaluation protocol that measures what an agent reads by editing the observation and the action list separately and by perturbing episodes in ways a routine cannot absorb. Specifically, applied to six open models, it shows that the strongest still solves 75\% of unseen games with every observation emptied, almost never retries an action that the environment reports as failed, and loses 27 points of success when told a false location.
\end{abstract}

\section{Introduction}

Recently, there has been a trend of training open language models as interactive agents, with ALFWorld \citep{shridhar2021alfworld} as the principal yardstick. Supervised fine-tuning on expert trajectories raised 7B models to between 70 and 90\% success \citep{zeng2024agenttuning,song2024trial,xi2025agentgym}, and multi-turn reinforcement learning with step-level credit assignment reached 90.8\% with Qwen2.5-7B \citep{feng2025group}. Reading these numbers as progress rests on a premise: a high success rate requires closed-loop behavior, since to \emph{clean some bowl and put it in cabinet} an agent must look for the bowl, notice when it appears, and track the effect of each action. Recent analyses have questioned parts of this premise, finding that the reasoning of RL-trained agents becomes insensitive to their input \citep{wang2026ragen} and that agents keep acting on their original goal when the instruction is corrupted \citep{liu2026diagnosing}, yet the success rates of the same kind of agents remain high, which neither analysis explains.

However, one concern naturally arises that the benchmark may not require closed-loop behavior at all. Figure~\ref{fig:teaser}(a) tests this premise directly on the unseen task \emph{clean some bowl and put it in cabinet}. A policy that ranks receptacle classes by how often bowls lie in them in the training games visits cabinets 6, 5, and 4 in that order, reads one fact at each stop, whether a bowl is there, and after the third stop executes \emph{take}, \emph{clean}, and \emph{place}, finishing in ten steps. This policy, which we call Patrol~\icon{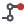}, solves all 134 unseen games, whereas the closed-loop expert that ships with the benchmark solves 90.3\% (Figure~\ref{fig:teaser}(c), dashed line). More importantly, even the one fact need not be observed. Figure~\ref{fig:teaser}(b) shows the last two stops as a trained agent receives them: the admissible-action list~\icon{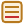} offers \texttt{take bowl 1 from cabinet 4} if and only if the observation shows the bowl, and Patrol reading only this list also solves every unseen game. ALFWorld success therefore certifies little beyond recognizing the target and following a search order, which is what an agent with input-insensitive reasoning can still do. The two channels can be told apart only by perturbing them: a decoy list that offers a \texttt{take} action for the target at every receptacle drops the list-reading Patrol to 18.7\% and leaves the observation-reading Patrol at 100\%, and silent failure of 30\% of the manipulations drops the latter to 33.6\% because it never checks what its actions did. A success rate measured on the unperturbed benchmark contains neither signal.

In this work, we present \oursname{}, an evaluation protocol that measures what an agent reads instead of whether it finishes. We rethink the evaluation of ALFWorld agents in two aspects: what success certifies, and through which channel the state reaches the agent. Concretely, \oursname{} has three components. First, scripted references bound how much observation the benchmark requires, because a test that a frequency-table routine passes cannot certify perception. Second, counterfactual probes at fixed expert states edit the observation, the admissible-action list, or both, in ways that change the correct next action, because only an edit confined to one channel can attribute a response to it. Third, closed-loop tests blind the observation, replace the list with a decoy, make actions fail silently, or give a true or false location hint, because responding to a local edit need not translate into success over an episode.

Our work exhibits several compelling facets. First, ALFWorld is solvable open-loop: Patrol solves 138 of the 140 seen games and all unseen games in 11.4 and 9.7 steps on average, above the benchmark's own expert on both splits. Second, the admissible-action list is a complete side channel for the decision-relevant state, and released RL agents read it: GiGPO-7B still solves 74.6\% of unseen games with every observation emptied, and the 1.5B agents respond more to edits of the list than of the observation. Third, the released agents barely track the outcome of their actions, although a single outcome check lifts Patrol under silent failures from 35.0 to 98.6\% on seen games: when a successful manipulation is reported as \texttt{Nothing happens.}, GiGPO-7B changes its next action with responsiveness 0.03, and a false location hint costs it 26.9 points. We release the scripts, the interventions, and 6{,}494 probes on 1{,}376 expert states, so that agents can be scored on what they read.

\section{Related Work}

\paragraph{LLM Agents.}
ALFWorld renders the ALFRED tasks \citep{shridhar2020alfred} as TextWorld games \citep{cote2019textworld} and is the default testbed for agents that interleave reasoning and acting \citep{yao2023react,shinn2023reflexion}. Supervised fine-tuning \citep{zeng2024agenttuning,chen2024agentflan,song2024trial,xi2025agentgym} and multi-turn reinforcement learning \citep{wang2025ragen,feng2025group,fei2025unleashing} have raised success steadily. These works report success rate and episode length; we ask what the benchmark requires to attain them.

\paragraph{Agent Grounding.}
The reasoning of RL-trained agents can become insensitive to their input \citep{wang2026ragen,wei2025gtr}, ReAct thoughts are often ungrounded \citep{kim2025reflact,verma2024brittle}, and agents ignore corrupted instructions, shuffled histories, and solutions planted in the environment \citep{liu2026diagnosing,liu2026history,englander2026agents}. Closest to us, \citet{yao2021reading} showed that a pre-LLM text-game agent given valid actions performs as well with hashed observations. We show that in ALFWorld the action list alone suffices for a perfect policy and give tests that separate the two channels for language agents.

\paragraph{Shortcut Learning.}
Policies cloned from observation histories can learn to copy past actions while ignoring the state \citep{dehaan2019causal,wen2020fighting}, an instance of shortcut learning \citep{geirhos2020shortcut}, and agentic benchmarks have been audited for such loopholes \citep{zhu2025establishing}. Methods that add transition-prediction losses to agent training \citep{li2026tapo,chen2026why} presume that agents under-use feedback; \oursname{} measures whether they do.

\section{\oursname{}}
\label{sec:method}

\begin{figure*}[t]
\centering
\includegraphics[width=\textwidth]{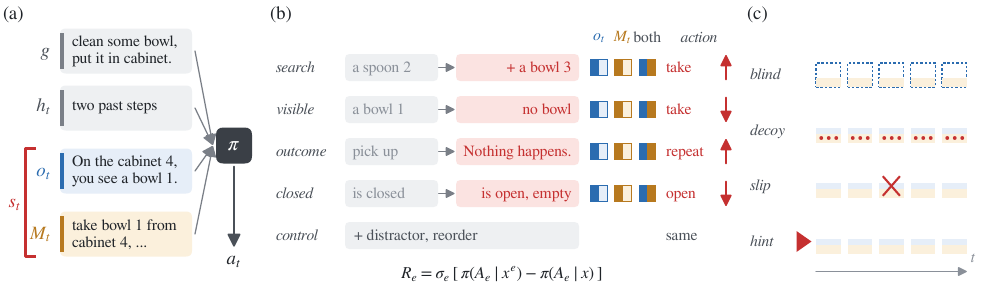}
\caption{\oursname{} tests through which channel an agent reads the state. (a) The agent input at one step. (b) Counterfactual probes that edit the observation, the action list, or both. (c) Closed-loop tests over an episode.}
\label{fig:method}
\end{figure*}

We first formalize the question over an agent's input channels (\S\ref{sec:problem}) and then describe the three components of \oursname{} (Figure~\ref{fig:method}): scripted references, counterfactual probes, and closed-loop tests (\S\ref{sec:protocol}).

\subsection{Problem Statement}
\label{sec:problem}

At step $t$ an agent receives $x_t=(g,h_t,o_t,M_t)$, the task, a window of past observation and action pairs, the current observation, and the admissible-action list, and samples an action $a_t\sim\pi(\cdot\mid x_t)$; this is the input under which the RL checkpoints of verl-agent are trained \citep{feng2025group}. The state $s_t$ reaches the policy twice (Figure~\ref{fig:method}(a)): $o_t$ describes it, and $M_t$ lists the actions it permits. Success rate is therefore silent on two questions: i) whether a policy uses $s_t$ beyond what a fixed routine needs, and ii) whether it reads $s_t$ from $o_t$ or from $M_t$.

\subsection{Protocol}
\label{sec:protocol}

\paragraph{Scripted References.}
Patrol visits the receptacles of a room in an order fixed by how often the target class lies in each receptacle class in the training games, opens what is closed, and at each stop reads one fact, whether an instance of the target is present, after which it runs the template of the task type (take; heat, cool, or clean at the appliance; place, or switch on a lamp). Patrol-$o$ reads the fact from $o_t$ and Patrol-$M$ from $M_t$; Patrol-$o$+verify also repeats an action answered with \texttt{Nothing happens.} We report the ALFWorld hand-coded expert, which tracks the state through privileged predicates, alongside.

\paragraph{Counterfactual Probes.}
We keep the states of the expert's solved validation episodes, so that every agent is probed on the same states and histories. Four kinds of state admit an edit that changes the correct action $A_e$ (Figure~\ref{fig:method}(b)): adding the target where the expert moves on (\emph{search}), removing it where the expert takes it (\emph{visible}), replacing the feedback of a successful manipulation with \texttt{Nothing happens.} (\emph{outcome}), and showing a closed receptacle as open and empty (\emph{closed}). Each edit is applied to $o_t$ alone, to $M_t$ alone, or to both consistently; adding an unrelated object or reordering the listed objects serves as a control. The responsiveness to an edit,
\begin{equation}
R_e=\sigma_e\big[\pi(A_e\mid x^{e})-\pi(A_e\mid x)\big],
\label{eq:resp}
\end{equation}
with $\sigma_e=\pm1$ for edits that should raise or lower $A_e$, is 1 when the policy follows the edit and 0 when it ignores it; a policy that reads the observation has $R^{o}\approx R^{oM}$, and one that reads the list has $R^{M}\approx R^{oM}$, whatever its success rate.

\paragraph{Closed-Loop Tests.}
These tests run complete episodes (Figure~\ref{fig:method}(c)): \emph{blind} empties every observation in the prompt and keeps the true list; \emph{decoy} adds to the list at every receptacle a \texttt{take} action for each instance of the target class, so that the list no longer reveals where the target is; \emph{slip} makes each accepted manipulation fail silently with probability 0.3, leaving the state unchanged \citep{tessler2019action}; and \emph{hint} appends to the task a receptacle that holds the target (true) or does not (false).

\paragraph{Agents and Evaluation Cost.}
We evaluate the 1.5B and 7B instruction-tuned Qwen2.5 models \citep{qwen2025qwen25} and the released verl-agent checkpoints trained from them on ALFWorld with PPO \citep{schulman2017ppo}, GRPO \citep{shao2024deepseekmath}, and GiGPO \citep{feng2025group} (1.5B) and with GiGPO (7B), in the prompt they were trained with (two-step history, greedy decoding). Scripts and closed-loop tests run on the 274 validation games (140 seen, 134 unseen) with the 50-step limit; the probe set contains 6{,}494 prompts on 1{,}376 expert states (400 search, 247 visible, 381 outcome, 348 closed), which an agent answers with eight samples each. With vLLM \citep{kwon2023vllm}, the whole protocol takes 15 to 40 minutes per agent on one H200 GPU.

\section{Results}
\label{sec:results}

\begin{table*}[t]
\centering
\footnotesize\setlength{\tabcolsep}{4.4pt}
\begin{tabular}{@{}lc>{\columncolor{oursl}}c>{\columncolor{oursl}}c>{\columncolor{oursl}}c>{\columncolor{oursl}}c>{\columncolor{oursl}}c>{\columncolor{oursl}}c@{\hspace{10pt}}cccc@{}}
\toprule
 & \multicolumn{7}{c}{Unseen success (\%)} & \multicolumn{4}{c}{Responsiveness to probes} \\
\cmidrule(lr){2-8}\cmidrule(l){9-12}
Policy & Std & Blind & Decoy & Both & Slip & Hint (T) & Hint (F) & $R^{o}$ & $R^{M}$ & $R^{oM}$ & $R^{oM}_{\text{out}}$ \\
\midrule
\multicolumn{12}{@{}l}{\textit{Scripted references}} \\
Expert (privileged) & 90.3 & -- & -- & -- & -- & -- & -- & -- & -- & -- & -- \\
Patrol-$o$ & 100.0 & -- & 100.0 & -- & \drop{33.6} & -- & -- & -- & -- & -- & -- \\
Patrol-$M$ & 100.0 & -- & \drop{18.7} & -- & -- & -- & -- & -- & -- & -- & -- \\
Patrol-$o$+verify & -- & -- & -- & -- & 100.0 & -- & -- & -- & -- & -- & -- \\
\midrule
\multicolumn{12}{@{}l}{\textit{Open models}} \\
Qwen2.5-1.5B-Instruct & 3.0 & 0.8 & 0.8 & 0.8 & 1.5 & 9.0 & 0.0 & 0.11 & 0.14 & 0.22 & 0.07 \\
\quad + PPO & 50.7 & 25.4 & 42.5 & 9.7 & 41.8 & 73.9 & 34.3 & 0.23 & 0.27 & 0.46 & 0.03 \\
\quad + GRPO & 60.5 & 18.7 & 49.3 & \drop{6.7} & 52.2 & 77.6 & 38.8 & 0.31 & 0.34 & 0.65 & 0.30 \\
\quad + GiGPO & 79.1 & 52.2 & 61.2 & \drop{25.4} & 67.2 & 91.8 & 61.9 & 0.24 & 0.44 & 0.72 & 0.08 \\
Qwen2.5-7B-Instruct & 18.7 & 9.7 & 19.4 & 8.2 & 10.4 & 36.6 & 3.0 & 0.23 & 0.12 & 0.32 & 0.01 \\
\quad + GiGPO & 91.8 & 74.6 & 83.6 & \drop{23.9} & 87.3 & 99.3 & 64.9 & 0.50 & 0.25 & 0.69 & 0.03 \\
\bottomrule
\end{tabular}
\caption{\oursname{} on the 134 unseen games. Shaded: closed-loop tests; Both blinds the observation and uses the decoy list; red: a drop of more than 50 points. $R^{o}$, $R^{M}$, $R^{oM}$: responsiveness (Eq.~\ref{eq:resp}) averaged over the four edit kinds, for edits to the observation, the list, or both; $R^{oM}_{\text{out}}$: the \emph{outcome} edit alone. Seen split: Appendix~\ref{app:agents}.}
\label{tab:main}
\end{table*}

\paragraph{Open-Loop Solvability.}
A fixed search order with one observed fact per receptacle suffices for ALFWorld. Patrol-$o$ solves every unseen game (Table~\ref{tab:main}) in 9.7 steps on average, and 138 of the 140 seen games (Appendix~\ref{app:scripts}), while the hand-coded expert, which tracks the state through privileged predicates, solves 90.3\% of unseen games. ALFWorld success therefore certifies that an agent can recognize the target when it is listed, together with a routine that a frequency table supplies.

\paragraph{Action-List Leakage.}
The action list alone carries the decision-relevant state. Patrol-$M$, which reads the list and ignores the observation, also solves every unseen game, and falls to 18.7\% when the list at every receptacle offers decoy \texttt{take} actions for every instance of the target class, while Patrol-$o$ is unaffected. Under silent failures, Patrol-$o$ falls to 33.6\% when each manipulation fails silently with probability 0.3, and a single check that repeats an action after \texttt{Nothing happens.} restores 100\% at under two extra steps.

\paragraph{Released Agents Read Both Channels.}
GiGPO-7B solves 91.8\% of unseen games, 74.6\% with every observation emptied and 83.6\% with the decoy list, but only 23.9\% when both channels are degraded, so either channel alone carries most of its success. The probes agree: GiGPO-7B responds more to edits of the observation ($R^{o}{=}0.50$) than of the list (0.25), whereas GiGPO-1.5B responds more to the list (0.44 vs.\ 0.24), and GRPO-1.5B keeps less than a third of its success without observations. Edits that should not matter change the actions of GiGPO-7B about as much as resampling does (total variation 0.21 vs.\ 0.20). Success rate thus ranks these agents by how often they finish, not by which input they act on.

\paragraph{Agents Barely Track Outcomes.}
If the feedback of a successful manipulation is replaced by \texttt{Nothing happens.}, every agent except GRPO-1.5B keeps its next action ($R^{oM}_{\text{out}}<0.1$), and GiGPO-7B repeats an action that silently failed in 13 of 403 cases. Its success under slips drops only to 87.3\% because it discovers the failure later, when the next manipulation is not admissible, at the cost of 7.4 extra steps per episode. A false location hint sends GiGPO-7B to the named receptacle first in every unseen game and costs it 26.9 points, so what it is told overrides its learned search order.

\section{Conclusion}
ALFWorld can be solved by a fixed search order and one observed fact per receptacle, and the admissible-action list given to trained agents supplies that fact on its own. Released RL agents read both channels, mostly ignore reports that an action failed, and follow what they are told. A success rate on ALFWorld should therefore come with tests that a routine cannot pass, such as those of \oursname{}.

\section*{Limitations}
Although \oursname{} separates the two input channels for six open models, each agent is evaluated with one greedy run per game in the prompt it was trained with, and agents trained without the action list or with supervised fine-tuning may read their input differently. The decoy list and the slips change the interface or the dynamics of the environment, so an agent may also fail them simply because such events never occur in training. Patrol's search order also reflects how similar object placements are across ALFWorld scenes, and benchmarks with less regular layouts would need a weaker reference. The next step is to fine-tune a control agent on perturbed games, which separates a lack of perception from a lack of exposure.

\section*{Broader Impact and Ethical Considerations}
Interactive benchmarks such as ALFWorld decide which training methods for language agents are adopted, and our findings change what a success rate on them can be taken to certify. Any benchmark that lists the admissible actions or places objects according to a learnable prior can be audited in the same way, which suggests reporting an open-loop scripted reference and an interface-perturbed score next to every success rate and designing agent inputs that do not duplicate the state. On the negative side, the tests could be used to tune agents that pass them without perceiving more, or to dismiss agents whose use of the action list is harmless in deployment. We therefore release the scripts and probes as diagnostics to be reported together with task success. All experiments use the public ALFWorld games and involve no human subjects, annotators, or personal data.

\bibliography{refs}

\begin{thebibliography}{31}
\providecommand{\natexlab}[1]{#1}

\bibitem[{Chen et~al.(2026)Chen, Zhu, Wang, Zhang, Wang, Zhou, Gao, Xiao
  et~al.}]{chen2026why}
Shiqi Chen, Tongyao Zhu, Zian Wang, Jinghan Zhang, Kangrui Wang, Ruochen Zhou,
  Siyang Gao, Teng Xiao, et~al. 2026.
\newblock Why do {LLM} agents fail in exploring new environments? {A}
  world-modeling perspective.
\newblock In \emph{Findings of the Association for Computational Linguistics:
  EMNLP 2026}.

\bibitem[{Chen et~al.(2024)Chen, Liu, Wang, Zhang, Liu, Lin, Chen, and
  Zhao}]{chen2024agentflan}
Zehui Chen, Kuikun Liu, Qiuchen Wang, Wenwei Zhang, Jiangning Liu, Dahua Lin,
  Kai Chen, and Feng Zhao. 2024.
\newblock {Agent-FLAN}: Designing data and methods of effective agent tuning
  for large language models.
\newblock In \emph{Findings of the Association for Computational Linguistics:
  ACL 2024}.

\bibitem[{C{\^o}t{\'e} et~al.(2019)C{\^o}t{\'e}, K{\'a}d{\'a}r, Yuan, Kybartas,
  Barnes, Fine, Moore, Hausknecht et~al.}]{cote2019textworld}
Marc-Alexandre C{\^o}t{\'e}, {\'A}kos K{\'a}d{\'a}r, Xingdi Yuan, Ben Kybartas,
  Tavian Barnes, Emery Fine, James Moore, Matthew Hausknecht, et~al. 2019.
\newblock {TextWorld}: A learning environment for text-based games.
\newblock In \emph{Computer Games: 7th Workshop, CGW 2018, Held in Conjunction
  with the 27th International Conference on Artificial Intelligence, IJCAI
  2018, Stockholm, Sweden, July 13, 2018, Revised Selected Papers}.

\bibitem[{de~Haan et~al.(2019)de~Haan, Jayaraman, and
  Levine}]{dehaan2019causal}
Pim de~Haan, Dinesh Jayaraman, and Sergey Levine. 2019.
\newblock Causal confusion in imitation learning.
\newblock In \emph{Advances in Neural Information Processing Systems}.

\bibitem[{Engl{\"a}nder et~al.(2026)Engl{\"a}nder, Althammer, {\"U}st{\"u}n,
  Gall{\'e}, and Sherborne}]{englander2026agents}
Leon Engl{\"a}nder, Sophia Althammer, Ahmet {\"U}st{\"u}n, Matthias Gall{\'e},
  and Tom Sherborne. 2026.
\newblock Agents explore but agents ignore: {LLMs} lack environmental
  curiosity.
\newblock \emph{arXiv preprint arXiv:2604.17609}.

\bibitem[{Fei et~al.(2025)Fei, Ji, Wang, Shi, Gong, and
  Qiu}]{fei2025unleashing}
Zhaoye Fei, Li~Ji, Siyin Wang, Junhao Shi, Jingjing Gong, and Xipeng Qiu. 2025.
\newblock Unleashing embodied task planning ability in {LLMs} via reinforcement
  learning.
\newblock \emph{arXiv preprint arXiv:2506.23127}.

\bibitem[{Feng et~al.(2025)Feng, Xue, Liu, and An}]{feng2025group}
Lang Feng, Zhenghai Xue, Tingcong Liu, and Bo~An. 2025.
\newblock Group-in-group policy optimization for {LLM} agent training.
\newblock In \emph{Advances in Neural Information Processing Systems}.

\bibitem[{Geirhos et~al.(2020)Geirhos, Jacobsen, Michaelis, Zemel, Brendel,
  Bethge, and Wichmann}]{geirhos2020shortcut}
Robert Geirhos, J{\"o}rn-Henrik Jacobsen, Claudio Michaelis, Richard Zemel,
  Wieland Brendel, Matthias Bethge, and Felix~A. Wichmann. 2020.
\newblock Shortcut learning in deep neural networks.
\newblock \emph{Nature Machine Intelligence}.

\bibitem[{Kim et~al.(2025)Kim, Rhee, Kim, Kim, Lee, Sung, and
  Jung}]{kim2025reflact}
Jeonghye Kim, Sojeong Rhee, Minbeom Kim, Dohyung Kim, Sangmook Lee, Youngchul
  Sung, and Kyomin Jung. 2025.
\newblock {ReflAct}: World-grounded decision making in {LLM} agents via
  goal-state reflection.
\newblock In \emph{Proceedings of the 2025 Conference on Empirical Methods in
  Natural Language Processing}.

\bibitem[{Kwon et~al.(2023)Kwon, Li, Zhuang, Sheng, Zheng, Yu, Gonzalez, Zhang,
  and Stoica}]{kwon2023vllm}
Woosuk Kwon, Zhuohan Li, Siyuan Zhuang, Ying Sheng, Lianmin Zheng, Cody~Hao Yu,
  Joseph~E. Gonzalez, Hao Zhang, and Ion Stoica. 2023.
\newblock Efficient memory management for large language model serving with
  {PagedAttention}.
\newblock In \emph{Proceedings of the 29th Symposium on Operating Systems
  Principles}.

\bibitem[{Li et~al.(2026)Li, Peng, Zhao, Hu, Liu, Su, and Li}]{li2026tapo}
Cong Li, Peixi Peng, Yisen Zhao, Xinyu Hu, Shudong Liu, Zhan Su, and Zhuojian
  Li. 2026.
\newblock {TAPO}: Transition-aware policy optimization for {LLM} agents.
\newblock \emph{arXiv preprint arXiv:2607.27973}.

\bibitem[{Liu et~al.(2026{\natexlab{a}})Liu, Lin, Wu, Chen, Yu, and
  Liu}]{liu2026diagnosing}
Jingyu Liu, Jiaen Lin, Xiaopeng Wu, Kehan Chen, Chuan Yu, and Yong Liu.
  2026{\natexlab{a}}.
\newblock Diagnosing task insensitivity in language agents.
\newblock \emph{arXiv preprint arXiv:2606.26918}.

\bibitem[{Liu et~al.(2026{\natexlab{b}})Liu, Wang, Jing, Zhou, and
  Liu}]{liu2026history}
Jingyu Liu, Zhiwen Wang, Yuxin Jing, Huanyu Zhou, and Yong Liu.
  2026{\natexlab{b}}.
\newblock When history fails to become experience: Action calibration in
  language agents.
\newblock \emph{arXiv preprint arXiv:2610.02769}.

\bibitem[{Schulman et~al.(2017)Schulman, Wolski, Dhariwal, Radford, and
  Klimov}]{schulman2017ppo}
John Schulman, Filip Wolski, Prafulla Dhariwal, Alec Radford, and Oleg Klimov.
  2017.
\newblock Proximal policy optimization algorithms.
\newblock \emph{arXiv preprint arXiv:1707.06347}.

\bibitem[{Shao et~al.(2024)Shao, Wang, Zhu, Xu, Song, Bi, Zhang, Zhang
  et~al.}]{shao2024deepseekmath}
Zhihong Shao, Peiyi Wang, Qihao Zhu, Runxin Xu, Junxiao Song, Xiao Bi, Haowei
  Zhang, Mingchuan Zhang, et~al. 2024.
\newblock {DeepSeekMath}: Pushing the limits of mathematical reasoning in open
  language models.
\newblock \emph{arXiv preprint arXiv:2402.03300}.

\bibitem[{Shinn et~al.(2023)Shinn, Cassano, Gopinath, Narasimhan, and
  Yao}]{shinn2023reflexion}
Noah Shinn, Federico Cassano, Ashwin Gopinath, Karthik Narasimhan, and Shunyu
  Yao. 2023.
\newblock Reflexion: Language agents with verbal reinforcement learning.
\newblock In \emph{Advances in Neural Information Processing Systems}.

\bibitem[{Shridhar et~al.(2020)Shridhar, Thomason, Gordon, Bisk, Han, Mottaghi,
  Zettlemoyer, and Fox}]{shridhar2020alfred}
Mohit Shridhar, Jesse Thomason, Daniel Gordon, Yonatan Bisk, Winson Han,
  Roozbeh Mottaghi, Luke Zettlemoyer, and Dieter Fox. 2020.
\newblock {ALFRED}: A benchmark for interpreting grounded instructions for
  everyday tasks.
\newblock In \emph{Proceedings of the IEEE/CVF Conference on Computer Vision
  and Pattern Recognition}.

\bibitem[{Shridhar et~al.(2021)Shridhar, Yuan, C{\^o}t{\'e}, Bisk, Trischler,
  and Hausknecht}]{shridhar2021alfworld}
Mohit Shridhar, Xingdi Yuan, Marc-Alexandre C{\^o}t{\'e}, Yonatan Bisk, Adam
  Trischler, and Matthew Hausknecht. 2021.
\newblock {ALFWorld}: Aligning text and embodied environments for interactive
  learning.
\newblock In \emph{International Conference on Learning Representations}.

\bibitem[{Song et~al.(2024)Song, Yin, Yue, Huang, Li, and Lin}]{song2024trial}
Yifan Song, Da~Yin, Xiang Yue, Jie Huang, Sujian Li, and Bill~Yuchen Lin. 2024.
\newblock Trial and error: Exploration-based trajectory optimization of {LLM}
  agents.
\newblock In \emph{Proceedings of the 62nd Annual Meeting of the Association
  for Computational Linguistics (Volume 1: Long Papers)}.

\bibitem[{Tessler et~al.(2019)Tessler, Efroni, and Mannor}]{tessler2019action}
Chen Tessler, Yonathan Efroni, and Shie Mannor. 2019.
\newblock Action robust reinforcement learning and applications in continuous
  control.
\newblock In \emph{International Conference on Machine Learning}.

\bibitem[{Verma et~al.(2024)Verma, Bhambri, and Kambhampati}]{verma2024brittle}
Mudit Verma, Siddhant Bhambri, and Subbarao Kambhampati. 2024.
\newblock On the brittle foundations of {ReAct} prompting for agentic large
  language models.
\newblock \emph{arXiv preprint arXiv:2405.13966}.

\bibitem[{Wang et~al.(2026)Wang, Gui, Jin, Wang, Liu, Wang, Chen, Li
  et~al.}]{wang2026ragen}
Zihan Wang, Chi Gui, Xing Jin, Qineng Wang, Licheng Liu, Kangrui Wang, Shiqi
  Chen, Linjie Li, et~al. 2026.
\newblock {RAGEN-2}: Reasoning collapse in agentic {RL}.
\newblock In \emph{International Conference on Machine Learning}.

\bibitem[{Wang et~al.(2025)Wang, Wang, Wang, Zhang, Li, Yang, Jin, Yu
  et~al.}]{wang2025ragen}
Zihan Wang, Kangrui Wang, Qineng Wang, Pingyue Zhang, Linjie Li, Zhengyuan
  Yang, Xing Jin, Kefan Yu, et~al. 2025.
\newblock {RAGEN}: Understanding self-evolution in {LLM} agents via multi-turn
  reinforcement learning.
\newblock \emph{arXiv preprint arXiv:2504.20073}.

\bibitem[{Wei et~al.(2025)Wei, Yang, Xing, Shi, Lu, and Ye}]{wei2025gtr}
Tong Wei, Yijun Yang, Junliang Xing, Yuanchun Shi, Zongqing Lu, and Deheng Ye.
  2025.
\newblock {GTR}: Guided thought reinforcement prevents thought collapse in
  {RL}-based {VLM} agent training.
\newblock In \emph{Proceedings of the IEEE/CVF International Conference on
  Computer Vision}.

\bibitem[{Wen et~al.(2020)Wen, Lin, Darrell, Jayaraman, and
  Gao}]{wen2020fighting}
Chuan Wen, Jierui Lin, Trevor Darrell, Dinesh Jayaraman, and Yang Gao. 2020.
\newblock Fighting copycat agents in behavioral cloning from observation
  histories.
\newblock In \emph{Advances in Neural Information Processing Systems}.

\bibitem[{Xi et~al.(2025)Xi, Ding, Chen, Hong, Guo, Wang, Guo, Yang
  et~al.}]{xi2025agentgym}
Zhiheng Xi, Yiwen Ding, Wenxiang Chen, Boyang Hong, Honglin Guo, Junzhe Wang,
  Xin Guo, Dingwen Yang, et~al. 2025.
\newblock {AgentGym}: Evaluating and training large language model-based agents
  across diverse environments.
\newblock In \emph{Proceedings of the 63rd Annual Meeting of the Association
  for Computational Linguistics (Volume 1: Long Papers)}.

\bibitem[{Yang et~al.(2024)Yang, Yang, Zhang, Hui, Zheng, Yu, Li, Liu
  et~al.}]{qwen2025qwen25}
An~Yang, Baosong Yang, Beichen Zhang, Binyuan Hui, Bo~Zheng, Bowen Yu,
  Chengyuan Li, Dayiheng Liu, et~al. 2024.
\newblock {Qwen2.5} technical report.
\newblock \emph{arXiv preprint arXiv:2412.15115}.

\bibitem[{Yao et~al.(2021)Yao, Narasimhan, and Hausknecht}]{yao2021reading}
Shunyu Yao, Karthik Narasimhan, and Matthew Hausknecht. 2021.
\newblock Reading and acting while blindfolded: The need for semantics in text
  game agents.
\newblock In \emph{Proceedings of the 2021 Conference of the North American
  Chapter of the Association for Computational Linguistics: Human Language
  Technologies}.

\bibitem[{Yao et~al.(2023)Yao, Zhao, Yu, Du, Shafran, Narasimhan, and
  Cao}]{yao2023react}
Shunyu Yao, Jeffrey Zhao, Dian Yu, Nan Du, Izhak Shafran, Karthik Narasimhan,
  and Yuan Cao. 2023.
\newblock {ReAct}: Synergizing reasoning and acting in language models.
\newblock In \emph{International Conference on Learning Representations}.

\bibitem[{Zeng et~al.(2024)Zeng, Liu, Lu, Wang, Liu, Dong, and
  Tang}]{zeng2024agenttuning}
Aohan Zeng, Mingdao Liu, Rui Lu, Bowen Wang, Xiao Liu, Yuxiao Dong, and Jie
  Tang. 2024.
\newblock {AgentTuning}: Enabling generalized agent abilities for {LLMs}.
\newblock In \emph{Findings of the Association for Computational Linguistics:
  ACL 2024}.

\bibitem[{Zhu et~al.(2025)Zhu, Jin, Pruksachatkun, Zhang, Liu, Cui, Kapoor,
  Longpre et~al.}]{zhu2025establishing}
Yuxuan Zhu, Tengjun Jin, Yada Pruksachatkun, Andy Zhang, Shu Liu, Sasha Cui,
  Sayash Kapoor, Shayne Longpre, et~al. 2025.
\newblock Establishing best practices in building rigorous agentic benchmarks.
\newblock In \emph{Advances in Neural Information Processing Systems, Datasets
  and Benchmarks Track}.

\end{thebibliography}

\appendix
\section{Probe Construction}
\label{app:probes}

Probe states come from the validation episodes that the hand-coded expert solves. A state is a \emph{search} state when the agent has just arrived at or opened a receptacle that does not hold the target class and the expert moves on, a \emph{visible} state when the receptacle holds the target and the expert takes it, an \emph{outcome} state when the previous take, heat, cool, or clean succeeded, and a \emph{closed} state when the agent arrives at a closed receptacle and the expert opens it. We keep at most three states per game and at most 400 states per kind.

Observation edits keep the ALFWorld surface form: objects are listed as ``a \emph{class} \emph{id}'' separated by commas with a final ``, and'', an empty receptacle reads ``you see nothing'', and the action list is kept in the alphabetical order the environment uses. An added target instance receives the smallest id of its class that does not occur in the state, so that the edit never refers to an object the agent has already handled, and a list edit adds or removes only the corresponding \texttt{take} action, so that all other entries are unchanged. The distractor class is drawn from 22 portable classes, excluding the task's object, receptacle, and lamp classes and any class that the state already contains.

\section{Scripted Policies on Both Splits}
\label{app:scripts}

\begin{table}[h]
\centering
\footnotesize\setlength{\tabcolsep}{3.4pt}
\begin{tabular}{@{}lccccr@{}}
\toprule
 & \multicolumn{2}{c}{Seen (140)} & \multicolumn{3}{c}{Unseen (134)} \\
\cmidrule(lr){2-3}\cmidrule(l){4-6}
Policy & SR & Steps & SR & Steps & $\Delta$ \\
\midrule
\multicolumn{6}{@{}l}{\textit{Standard interface}} \\
Expert (privileged) & 83.6 & -- & 90.3 & -- & \\
Patrol-$o$ & 98.6 & 11.4 & 100.0 & 9.7 & \\
Patrol-$M$ & 97.1 & 11.9 & 100.0 & 9.7 & \\
\midrule
\multicolumn{6}{@{}l}{\textit{Decoy list}} \\
\rowcolor{oursl} Patrol-$o$ & 98.6 & 11.4 & 100.0 & 9.7 & 0.0 \\
\rowcolor{oursl} Patrol-$M$ & \drop{17.1} & 42.3 & \drop{18.7} & 41.7 & $-$81.3 \\
\midrule
\multicolumn{6}{@{}l}{\textit{Silent slips} ($p{=}0.3$)} \\
\rowcolor{oursl} Patrol-$o$ & \drop{35.0} & 36.6 & \drop{33.6} & 36.5 & $-$66.4 \\
\rowcolor{oursl} Patrol-$o$+verify & 98.6 & 13.3 & 100.0 & 11.4 & 0.0 \\
\bottomrule
\end{tabular}
\caption{Success (\%) and mean episode length of scripted policies on the validation games; scripts are deterministic. Shaded: \oursname{} tests. $\Delta$: change in unseen success from the standard interface. Red: a drop of more than 50 points in success rate.}
\label{tab:scripts}
\end{table}

Both seen failures of Patrol-$o$ exhaust the step limit (Table~\ref{tab:scripts}): in one the ladle lies on the dining table that is also its destination, ranked by the prior after 27 cabinets; in the other the remaining CDs lie on a desk that shares a location with the drawer where the script stands, so \texttt{go to desk 1} returns \texttt{Nothing happens.}, and the desk is never inspected by the script.

\section{Released Agents}
\label{app:agents}

We use these Hugging Face Hub checkpoints, with \texttt{X} one of \texttt{ppo}, \texttt{grpo}, and \texttt{gigpo}:
{\small\begin{itemize}\setlength\itemsep{0pt}
\item \texttt{langfeng01/}\\\texttt{GiGPO-Qwen2.5-7B-Instruct-ALFWorld}
\item \texttt{chanyoungkim/}\\\texttt{X-qwen2.5-1.5b-alfworld-step140}
\end{itemize}}
\noindent Each agent sees the verl-agent prompt with a history window of two steps and samples its action greedily (temperature 1.0 and eight samples for the probes); an output without a parsable \texttt{<action>} tag is executed as an invalid command, as in verl-agent. Table~\ref{tab:agents_seen} gives the seen-split results. When the action list is removed from the prompt, every agent solves no game: GiGPO-7B, for example, issues \texttt{go to coffee table 1} for the receptacle \texttt{coffeetable 1}, so the list also supplies the names of the actions, and we do not use this condition to measure perception.

\begin{table}[h]
\centering
\footnotesize\setlength{\tabcolsep}{2.5pt}
\begin{tabular}{@{}lccccccc@{}}
\toprule
Agent & Std & Blind & Decoy & Both & Slip & H(T) & H(F) \\
\midrule
Qwen2.5-1.5B & 4.3 & 2.9 & 4.3 & 0.7 & 4.3 & 13.6 & 1.4 \\
\quad + PPO & 58.6 & 23.6 & 49.3 & 15.0 & 53.6 & 79.3 & 35.7 \\
\quad + GRPO & 64.3 & 22.9 & 49.3 & 14.3 & 58.6 & 82.9 & 43.6 \\
\quad + GiGPO & 75.7 & 50.0 & 65.0 & 27.9 & 69.3 & 86.4 & 72.9 \\
Qwen2.5-7B & 17.1 & 10.7 & 20.7 & 6.4 & 12.9 & 35.7 & 4.3 \\
\quad + GiGPO & 97.1 & 78.6 & 89.3 & 35.0 & 93.6 & 97.1 & 71.4 \\
\bottomrule
\end{tabular}
\caption{Success (\%) of the released agents on the 140 seen games, under the conditions of Table~\ref{tab:main}.}
\label{tab:agents_seen}
\end{table}

\section{Task-Type Breakdown}
\label{app:tasktype}

\begin{table}[h]
\centering
\footnotesize\setlength{\tabcolsep}{1.75pt}
\begin{tabular}{@{}lccccccc@{}}
\toprule
 & Pick & Look & Clean & Heat & Cool & Two & All \\
Patrol & (24) & (18) & (31) & (23) & (21) & (17) & (134) \\
\midrule
$o$ & 100.0 & 100.0 & 100.0 & 100.0 & 100.0 & 100.0 & 100.0 \\
$M$ & 100.0 & 100.0 & 100.0 & 100.0 & 100.0 & 100.0 & 100.0 \\
\midrule
\rowcolor{oursl} $M$, decoy & \drop{12.5} & \drop{38.9} & \drop{32.3} & \drop{0.0} & \drop{23.8} & \drop{0.0} & \drop{18.7} \\
\rowcolor{oursl} $o$, slips & \drop{33.3} & \drop{44.4} & \drop{29.0} & \drop{39.1} & \drop{38.1} & \drop{17.6} & \drop{33.6} \\
\rowcolor{oursl} $o$+verify, slips & 100.0 & 100.0 & 100.0 & 100.0 & 100.0 & 100.0 & 100.0 \\
\bottomrule
\end{tabular}
\caption{Success (\%) of Patrol on the unseen games by task type, with the number of games in parentheses. Shading and red as in Table~\ref{tab:scripts}.}
\label{tab:tasktype}
\end{table}

Both Patrol variants solve every task type, so no type of ALFWorld task requires more than the one fact per stop (Table~\ref{tab:tasktype}). The decoy list hurts most where the template needs the target in hand before anything else can happen (heat and pick-two tasks, 0.0\%), and slips hurt most on pick-two tasks (17.6\%), which contain the most manipulations; an outcome check removes the loss on every type.

\end{document}